\documentclass{article} 
\usepackage{iclr2021_conference,times}

\usepackage{amsmath,amsfonts,bm}

\def\eqref#1{equation~\ref{#1}}

\def\1{\bm{1}}

\DeclareMathAlphabet{\mathsfit}{\encodingdefault}{\sfdefault}{m}{sl}
\SetMathAlphabet{\mathsfit}{bold}{\encodingdefault}{\sfdefault}{bx}{n}

\usepackage{graphicx}  
\graphicspath{images}
\usepackage{wrapfig}
\usepackage{caption}
\usepackage{subcaption}
\usepackage{colortbl}
\usepackage{xcolor}
\usepackage{selectp} 
\usepackage{floatrow} 
\definecolor{darkgray}{rgb}{0.25, 0.25, 0.25}
\newcommand{\rng}[2]{\textcolor{darkgray}{\textsubscript{[#1 - #2]}}}
\newcommand{\e}[1]{$\times 10^{#1}$}
\newcolumntype{C}{>{\centering\arraybackslash}p{0.08cm}}

\usepackage{physics}
\usepackage{amsmath}
\usepackage{tikz}
\usepackage{mathdots}
\usepackage{yhmath}
\usepackage{cancel}
\usepackage{color}
\usepackage{siunitx}
\usepackage{array}
\newcolumntype{G}{>{\color{darkgray}}c}
\usepackage{multirow}
\usepackage{amssymb}
\usepackage{gensymb}
\usepackage{tabularx}
\usepackage{booktabs}
\usepackage{enumitem} 
\usetikzlibrary{fadings}
\usetikzlibrary{patterns}
\usetikzlibrary{shadows.blur}
\usetikzlibrary{shapes}

\usepackage{booktabs}
\usepackage{multirow}
\usepackage{hyperref}
\usepackage{url}

\title{Deep Ensembles for Low-Data\\ Transfer Learning}

\author{%
  Basil Mustafa,  Carlos Riquelme,  Joan Puigcerver, \\
  \textbf{Andr\'e Susano Pinto,  Daniel Keysers,  Neil Houlsby} \\
  
  Google Research 
  
}

\iclrfinalcopy 
\begin{document}

\maketitle

\begin{abstract}
In the low-data regime, it is difficult to train good supervised models from scratch. Instead practitioners turn to pre-trained models, leveraging transfer learning.
Ensembling is an empirically and theoretically appealing way to construct powerful predictive models, but the predominant approach of training multiple deep networks with different random initialisations collides with the need for transfer via pre-trained weights.
In this work, we study different ways of creating ensembles from pre-trained models. We show that the nature of pre-training itself is a performant source of 
diversity, and propose a practical algorithm that efficiently identifies a subset of pre-trained models for any downstream dataset.
The approach is simple: Use nearest-neighbour 
accuracy to rank pre-trained models, fine-tune the best ones with a small hyperparameter sweep, and greedily construct an ensemble to minimise validation cross-entropy.
When evaluated together with strong baselines on 19 different downstream tasks (the Visual Task Adaptation Benchmark), this achieves state-of-the-art performance at a much lower inference budget, even when selecting from over 2,000 pre-trained models.
We also assess our ensembles on ImageNet variants and show improved robustness to distribution shift.

\end{abstract}
\section{Introduction}
There are many ways to construct models with minimal data. It has been shown that fine-tuning pre-trained deep models is a compellingly simple and performant approach \citep{dhillon2019baseline, alex2019big}, and this is the paradigm our work operates in.
It is common to use networks pre-trained on ImageNet~\citep{deng2009imagenet}, but 
recent works show considerable improvements by careful, task-specific pre-trained model selection \citep{ngiam2018domain,puigcerver2020experts}.



Ensembling multiple models is a powerful idea that often leads to better predictive performance.
Its secret relies on combining different predictions.
The source of diversity for deep networks has been studied \citep{fort2019deep,wenzel2020hyperparameter}, though not thoroughly in the low-data regime.
Two of the most common approaches involve training independent models from scratch with (a)~different random initialisations, (b)~different random subsets of the training data.
Neither of these are directly applicable downstream with minimal data, as we require a pre-trained initialisation to train competitive models\footnote{For an illustration of the importance of using pre-trained models in the low-data regime see Appendix~\ref{app:from_scratch}.}, and data scarcity makes further data fragmentation impractical.
We study some ways of encouraging model diversity in a supervised transfer-learning setup, but fundamentally argue that the nature of pre-training is itself an easily accessible and valuable form of diversity.

Previous works consider the construction of ensembles from a set of candidate models \citep{caruana2004ensembleselection}. Services such as Tensorflow Hub \citep{tfhub} and PyTorch Hub \citep{pytorch-hub} contain hundreds of pre-trained models for computer vision; these could all be fine-tuned on a new task to generate candidates. Factoring in the cost of hyperparameter search, this may be prohibitively expensive. We would like to know how suited a pre-trained model is for our given task \emph{before} training it. This need has given rise to cheap proxy metrics which assess this suitability \citep{puigcerver2020experts}. We use such metrics - leave-one-out nearest-neighbour ($k$NN) accuracy, in particular - as a way of selecting a \textit{subset} of pre-trained models, suitable for creating diverse ensembles of task-specific experts. We show that our approach is capable of quickly narrowing large pools (up to 2,000) of candidate pre-trained models down to manageable (15 models) task-specific sets, yielding a practical algorithm in the common context of the availability of many pre-trained models.

\begin{figure}
    \centering
    \vspace{-1ex}
    \includegraphics[width=0.95\textwidth]{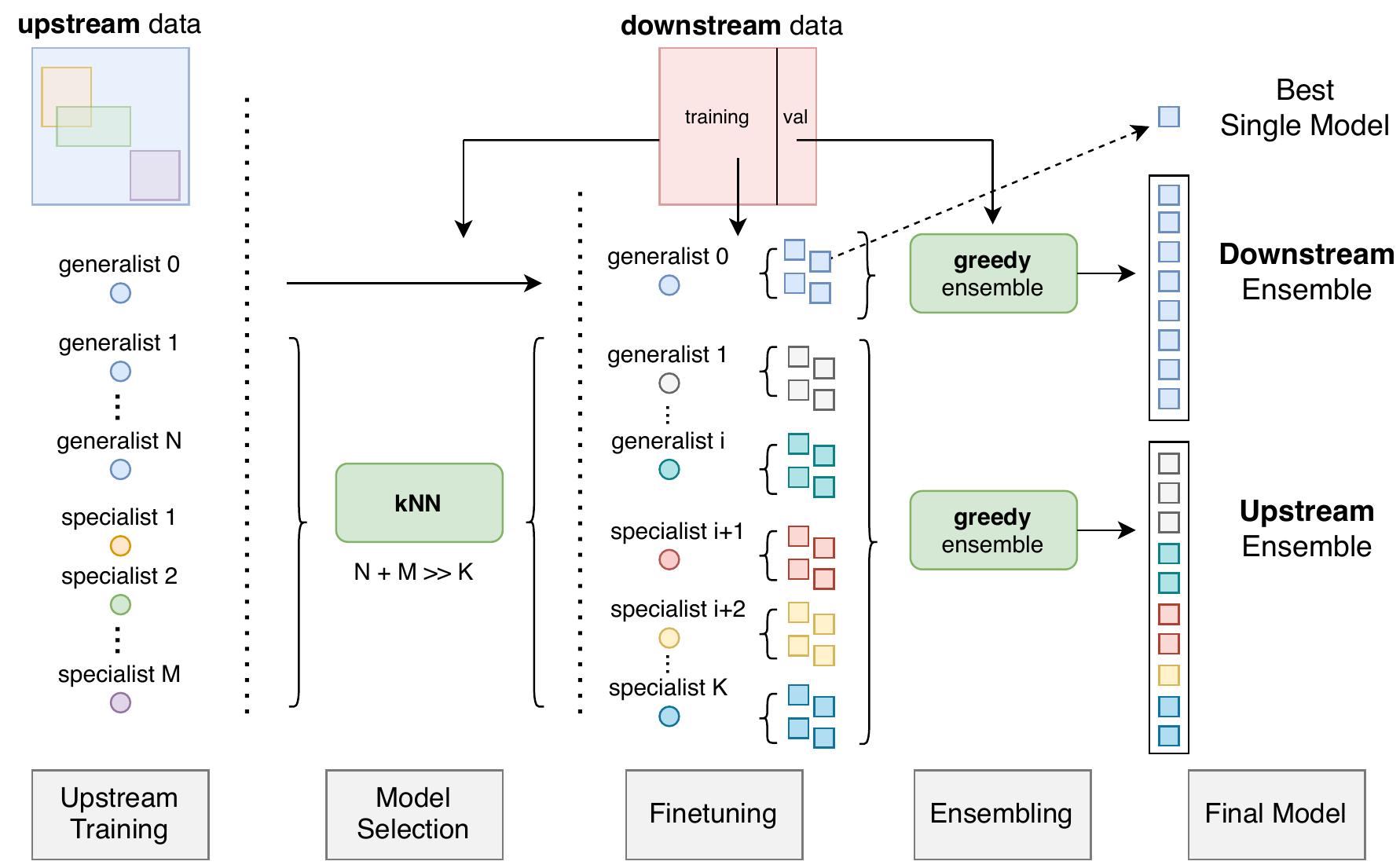}
     \caption{Overview of the different ways of constructing diverse ensembles studied in this work. We propose an algorithm that
     exploits diversity in a large pool of pre-trained models, by using leave-one-out $k$-nearest-neighbour ($k$NN) accuracy to select a subset to form the ensemble.}
    \label{fig:algorithm}
\end{figure}

We first experiment with sources of downstream diversity (induced only by hyperparameterisation, augmentation or random data ordering), giving significant performance boosts over single models. Using our algorithm on different pools of candidate pre-trained models, we show that various forms of upstream diversity produce ensembles that are more accurate and robust to domain shift than this. Figure~\ref{fig:algorithm} illustrates the different approaches studied in our work. Ultimately, this new form of diversity improves on the Visual Task Adaptation Benchmark \citep{zhai2019largescale} SOTA by 1.8\%.


The contributions of this paper can be summarized as follows:
\begin{itemize}[leftmargin=*] 
     \item We study ensembling in the context of transfer learning in the low data regime \& propose a number of ways to induce advantageous ensemble diversity which best leverage pre-trained models.
     \item We show that diversity from upstream pre-training achieves better accuracy than that from the downstream fine-tuning stage (+1.2 absolute points on average across the 19 downstream classification VTAB tasks), and that it is more robust to distribution shift (+2.2 absolute average accuracy increase on distribution shifted ImageNet variants).
     \item We show that they also surpass the accuracy of large SOTA models (76.2\% vs. 77.6\%) at a much lower inference cost, and achieve equal performance with less than a sixth of the FLOPS.
     \item We extend the work from \citet{puigcerver2020experts} and demonstrate the efficacy of $k$NN accuracy as a cheap proxy metric for selecting a \emph{subset} of candidate pre-trained models.
\end{itemize}


\section{Creating ensembles from pre-trained models}
We first formally introduce the technical problem we address in this paper.
Next we discuss baseline approaches which use a single pre-trained model, and then we present our method that exploits using multiple pre-trained models as a source of diversity.

\subsection{The Learning Setup: Upstream, Model Selection, Downstream}
Transfer learning studies how models trained in one context boost learning in a different one.
The most common approach pre-trains a single model on a large dataset such as ImageNet, to then tune the model weights to a downstream task.
Despite algorithmic simplicity, this idea has been very successful.
In a downstream low-data scenario, it is more difficult for a one-size-fits-all approach to triumph as 
specializing the initial representation becomes harder.
As in \citet{puigcerver2020experts}, we explore the scenario where a \emph{range} of pre-trained models is available, 
and we can look at the target data to make a decision on which models to fine-tune.
However, we generalize and improve it by simultaneously selected several models for fine-tuning,
since downstream tasks may benefit from combining expert representations aimed 
at capturing different aspects of the learning task: for instance, on a natural scenery dataset one could merge different models that focus on animals, plants, food, or buildings.
Fine-tuning all pre-trained models to pick the best one is a sensible strategy, but rarely feasible.
To keep the algorithms practical, we identify two compute budgets that should be controlled for:
The \emph{fine-tuning} budget, i.e. the total number of models we can fine-tune on a downstream task; and 
the \emph{inference} budget, the maximum size of the final model.


\subsection{Baselines: Diversity from downstream training}
The baselines we propose leverage transfer learning by requiring a pre-trained model - this is crucial, see Appendix~\ref{app:from_scratch}.
We use a strong \emph{generalist} model (BiT-ResNet 50s from \citet{alex2019big}, trained on all upstream data) and consider three methods to create a model set for ensemble selection.

\textbf{Random Seeds}. Fine-tuning a generalist model multiple times with \emph{fixed} hyperparameters will yield different classifiers, analagous to the DeepEnsembles of \citet{balaji2017ensembles}. Note, here we can only take advantage of randomised data ordering/augmentation, which \citet{fort2019deep} showed, though useful, was not as beneficial as diversity from random initalisation.

\textbf{HyperEnsembles}.
Hyperparameter diversity was recently shown to further improve DeepEnsembles~\citep{wenzel2020hyperparameter}.
We define a hyperparameter search space, randomly sample as many configurations as we have fine-tuning budget, and fine-tune the generalist on downstream data with each of those configurations. Further details on training are given in Appendix  \ref{app:training_hes}.

\textbf{AugEnsembles}.
We generate a set of models by fine-tuning the generalist on each task with randomly sampled \textit{families} of augmentation (but fixed hyperparameters). Details are in Appendix \ref{app:training_aes}.

\subsection{Our method: Diversity from upstream pre-training}
\citet{fort2019deep} explain the strong performance of classical ensembling approaches -- independently training randomly initialised deep networks -- by showing that each constituent model explores a different mode in the function space.
For transfer learning, \citet{neyshabur2020transferred} show that with pre-trained weights, fine-tuned models stay in a local `basin' in the loss landscape. Combining both gives a compelling reasoning for the use of multiple pre-trained networks for transfer with ensembles, as we propose here. Instead of diversity from downstream fine-tuning, we show that in the low data regime, better ensembles can be created using diversity from pre-training.

We consider three sources of upstream diversity.
First, we consider generalists that were \textit{pre-trained} with different random seeds on the same architecture and data.
Second, we consider \textit{experts}, specialist models which were pre-trained on different subsets of the 
large upstream dataset.
Lastly, we exploit diversity in scale -- pre-trained models with architectures of different sizes.
Given a pool of candidate models containing such diversity, we propose the following algorithm (Figure~\ref{fig:algorithm}):

\textbf{1. Pre-trained model selection.}
Fine-tuning all experts on the new task would be prohibitively expensive. Following \citet{puigcerver2020experts}, we rank all the models by their $k$NN leave-one-out accuracy as a proxy for final fine-tuned accuracy, instead keeping the $K$ best models (rather than 1).

\textbf{2. Fine-tuning.}
We add a fully connected layer to each model's final representation, and then train the whole model by minimising categorical cross-entropy via SGD.
Given a pool of $K$ pre-trained models from stage~1, we tune each with 4 learning rate schedules, yielding a total of $L = 4K$ models for the step 3 (Usually $K=15$ and $L=60$). See Appendix \ref{app:finetuning} for more details.

\textbf{3. Ensemble construction.}
This is shared among all presented ensembles. We use the greedy algorithm introduced by \cite{caruana2004ensembleselection}. At each step, we greedily pick the next model which minimises cross-entropy on the validation set when it is ensembled with already chosen models.

These steps are independently applied to each task; each step makes use of the downstream dataset, so each dataset gets a tailored set of pre-trained models to create the ensemble pool and therefore very different final ensembles result.
We also considered a greedy ensembling algorithm in $k$NN space which aims to sequentially pick complementary models which will likely ensemble well together (Appendix~\ref{app:greedy_NN}), but picking top-$K$ was generally better.

\subsubsection{Combined approaches}
\label{sec:combine}
The diversity induced by different upstream models and distinct downstream hyperparameters should be complementary. Given a fine-tuning budget of $L$, we can set the number of pre-trained models $K$ in advance, providing each of them with a random hyperparameter sweep of size ${L}/{K}$.
However, for some tasks it may be more beneficial to have fewer different pre-trained models and a wider sweep, or vice versa. We aim to dynamically set this balance per-dataset using a heuristic based on the $k$NN accuracies; namely, we keep all pre-trained models within some threshold percentage $\tau$\% of the top $k$NN accuracy, up to a maximum of $K = 15$. Ideally, this would adaptively discard experts poorly suited to a given task, whose inclusion would likely harm ensemble performance. The saved compute budget is then used to squeeze more performance from available experts by testing more hyperparameters, and hopefully leading to greater \emph{useful} diversity.
We arbitrarily set $\tau = 2\%$  for our experiments, but this choice could likely be improved upon. Appendix~\ref{app:num_experts_threshold} shows how the number of models picked varies per task.

\subsubsection{Pre-training models}
We use BiT ResNets pre-trained on two large upstream datasets with hierarchical label spaces: JFT-300M \citep{jft300m} and ImageNet-21k \citep{deng2009imagenet}.
We consider two types of pre-trained models.
\emph{Generalists} are trained on the entire upstream dataset. In particular, we consider 15 JFT ResNet-50 generalists that were pre-trained with different random initalisations.
\emph{Experts} are generated by splitting the hierarchical label spaces into sub-trees and training independent models on the examples in each sub-tree. We pre-train 244 experts from JFT and 50 from ImageNet21k, following the protocol of \citep{puigcerver2020experts} (see Appendix \ref{app:training_fes}).
For low-data downstream tasks, this is by far the most expensive stage of the process. It is however only incurred once, and its cost is amortized as new downstream tasks are served, since any downstream task can reuse them.
\section{Ensemble evaluation}
\label{sec:transfer}

\textbf{Downstream Tasks}.
We evaluate our models on the Visual Task Adaptation Benchmark \citep{zhai2019largescale}: 19 diverse downstream classification tasks, split into `natural', `specialised' and `structured' categories.
As we are primarily interested in low-data regimes, the tasks only have 1000 training datapoints (i.e., VTAB\textsubscript{1K}) with a number of classes ranging from 2 to 397.
We split data into 800 training examples and 200 validation examples.
See Appendix \ref{app:vtab} for more information.

\textbf{Test Performance}.
For our final competitive models, we first train all the individual models on the 800 training points.
Then, we use the 200 validation data points to find the best ensemble (both running the greedy algorithm and choosing the overall ensemble size). For the resultant ensemble, we retrain constituent models on the full 1000 data points, and evaluate it on the test data.


\textbf{Robustness}.
We train ExpertEnsembles and HyperEnsembles on ImageNet \citep{deng2009imagenet}. While ImageNet does not match our low-data regime of interest, previous work and additional datasets allow us to conveniently measure robustness and uncertainty metrics. 
Thus, alongside reporting the accuracy on the official validation split, we assess models on a number of ImageNet-based robustness benchmarks, aiming to quantify calibration and robustness to distribution shift \citep{djolonga2020robustness}.
More details on these variants are available in Appendices \ref{app:imagenet} and \ref{app:imagenet_training}.
\section{Experimental Results}
\label{sec:experiments}
Unless otherwise specified, all experiments use a fine-tuning budget and inference budget of 60 and 15 models respectively. This was set arbitrarily; we experiment with both budgets to see the effect. 

\begin{figure}[tb]
\centering
\begin{subfigure}{0.60\textwidth}
  \centering
  \includegraphics[height=4cm]{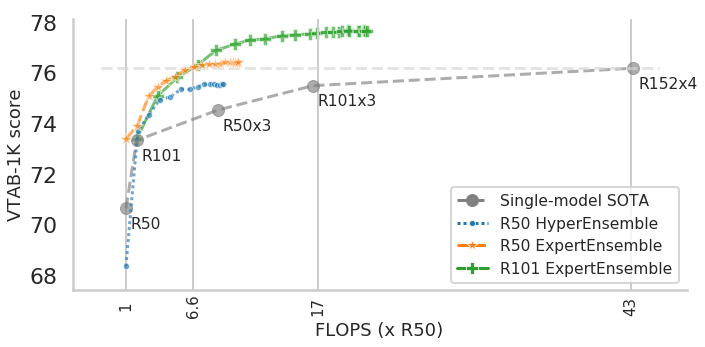}
  \caption{JFT pre-trained models.}
  \label{fig:flops_jft}
\end{subfigure}\hfill
\begin{subfigure}{.37\textwidth}
  \centering
  \includegraphics[height=4cm]{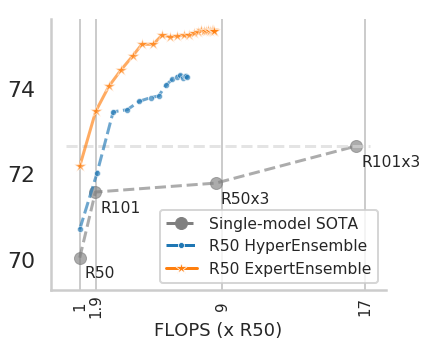}
  \caption{Imagenet21k pre-trained models.}
  \label{fig:flops_inet21k}
\end{subfigure}
\caption{Inference cost vs. VTAB\textsubscript{1K} performance. 
State-of-the-art generalist models of different scales are compared against ensembles with varying inference budgets.}
\label{fig:flops}
\end{figure}

\subsection{The Value of Ensembles}
We first show that ensembles in low-data transfer regimes dramatically beat their single-model counterparts --which are often much larger networks.
Figure~\ref{fig:flops} and Table~\ref{table:ensembles_vs_large_models} compare our best ensembles (which all use upstream or combined diversity) on VTAB\textsubscript{1K} tasks.
Our baselines are BiT models from \citep{alex2019big}, which had until now state-of-the-art performance.

\begin{table}[b]
\centering
\caption{Test accuracy of our best ensembles against reproduced baselines from \citet{alex2019big} [$^*$].
For each dataset, we take the median of three independent runs. Rows show the average over datasets.
Bootstrapped confidence intervals at the 95\% level. The source of diversity for ensembles is shown:
U = upstream (during pre-training) and C = combined (pre-training and fine-tuning). 
\label{table:ensembles_vs_large_models}}
\begin{tabular}{@{}lccGGG@{}}
\toprule
Description           & Diversity & VTAB\textsubscript{1K}  & Natural & Specialised & Structured \\
\midrule
\textbf{JFT} BiT-L R50$^*$          &--- & 70.6\rng{70.4}{71.0} & 77.8   & 83.6        & 57.9      \\
\textbf{JFT} BiT-L R152x4$^*$       &---   & 76.2\rng{74.5}{76.7} & 86.0   & 87.0       & 62.2      \\
\textbf{JFT} R50 Experts + Generalists & U & 76.8\rng{76.4}{77.0}  &  82.6  &  85.8  &  67.2    \\
\textbf{JFT} R101 Experts      & U & \textbf{77.6}\rng{77.4}{77.8} & 83.6   & 86.4        & 68.0      \\ 
\midrule 
\textbf{INet21k} BiT-M R50$^*$       &---    & 70.0\rng{69.6}{70.5} & 77.0   & 84.7        & 56.6      \\
\textbf{INet21k} BiT-M R101x3$^*$   &---    & 72.7\rng{72.1}{73.5} & 80.3   & 85.7        & 59.4      \\
\textbf{INet21k} R50 Experts       & U & 75.3\rng{74.5}{75.6} & 79.9   & 85.7        & 66.1      \\
\textbf{INet21k} R50 HyperExperts  & C & \textbf{75.6}\rng{74.8}{75.8} & 79.9   & 85.5        & 67.0      \\
\bottomrule
\end{tabular}
\end{table}

\textbf{JFT pre-trained models.} The most standard approach --fine-tuning a single R50 model trained on all of JFT-- leads to an average accuracy of 70.6\%. It greatly lags behind compared to the ensemble-based algorithms; in particular, the difference is striking for structured and natural datasets.
On average, the ensembles selected between 9 and 10 downstream models --this number greatly varies depending on the task, e.g. 3 were selected for CalTech101 and 14 for Diabetic Retinopathy.
Accordingly, capacity-wise, it makes sense to compare the ensembles to larger ResNets. 
Table~\ref{table:ensembles_vs_large_models} shows that the JFT R50 ensembles match or slightly beat the performance of a R152x4. In particular, the ensembles offer a large advantage in settings where tasks diverge from single-object recognition, e.g.\ in the structured datasets.
Even in natural domains, the experts have a better accuracy/FLOP ratio than the R152x4, which has $40\times$ more parameters than a single R50.
Even more significant is the difference in inference \emph{time}, as ensemble predictions can be easily parallelized.

\textbf{ImageNet21k pre-trained models.}
The story is fairly similar for the pool of ImageNet21k experts.
The ensembles select on average between 7 and 8 models --again, strongly task-dependent.
Ensembles' average performance improvement with respect to a single R50 trained on all of ImageNet21k is around 5 absolute points (more than 7\%).
We also consider a much larger generalist baseline, in this case a R101x3, which has more than $16\times$ as many parameters as a single R50. Still the ensembles far outperform it, especially in structured datasets.

\textbf{Overall.} A more complete story supporting the ensembles' 
value is depicted in Figure~\ref{fig:flops}.
The gray dashed line represents the previous Pareto frontier of VTAB\textsubscript{1K} average accuracy per FLOP. The ensemble models 
dominate the 
state-of-the-art, indicating their efficacy in the low-data regime.

\subsection{The Value of Upstream Diversity}
Results in Table~\ref{table:upstream_vs_downstream} suggest that upstream diversity improves downstream diversity. 
For both JFT and ImageNet21k pre-training, ensembles benefiting from upstream sources of diversity 
outperform their downstream-based counterparts. Combining both gives a further small boost for expert ensembles.

\begin{table}[tb]
\centering
\caption{\textbf{Upstream diversity gives better ensembles}. Test accuracy of different ensembles. 
For each dataset, we take the median of three independent runs. Rows show the average over datasets. 
Bootstrapped confidence intervals at the 95\% level.
The source of diversity is noted as: 
D = downstream (during fine-tuning), U = upstream (during pre-training) and C = combined (both).}
\label{table:upstream_vs_downstream}
\begin{tabular}{@{}lccGGG@{}}
\toprule
Description           & Diversity & VTAB\textsubscript{1K}  & Natural & Specialised & Structured \\ \midrule
\textbf{JFT} R50 Seeds  & D & 74.9\rng{74.5}{75.1}  &  78.5  &  85.9  &  66.2 \\
\textbf{JFT} R50 Augs   & D &  73.7\rng{73.4}{75.7}  &  80.6  &  86.8  &  61.2      \\
\textbf{JFT} R50 Hypers & D & 75.6\rng{75.2}{75.7} & 80.1   & 85.6       & 66.6      \\ \arrayrulecolor{gray}\midrule
\textbf{JFT} R50 Generalists+Experts & U &  76.8\rng{76.4}{77.0}  &  82.6  &  85.8  &  67.2     \\ 
\textbf{JFT} R50 Hyper(Gen. + Experts) & C & 77.1\rng{76.9}{77.4}  &  82.4  &  86  &  68.1 \\
\textbf{JFT} R50 Aug(Gen. + Experts) & C & \textbf{77.6} \rng{76.8}{77.9}  &  82.7  &  86.4  &  68.7 \\ \midrule
\arrayrulecolor{black} \midrule
\textbf{INet21k} R50 Hypers          & D & 74.2\rng{73.7}{74.8} & 79.6   & 86.2        & 63.5      \\
\textbf{INet21k} R50 Experts       & U & 75.3\rng{74.5}{75.6} & 79.9   & 85.7        & 66.1      \\
\textbf{INet21k} R50 HyperExperts  & C &\textbf{75.6}\rng{74.8}{75.8} & 79.9   & 85.5        & 67.0      \\
\bottomrule
\end{tabular}
\end{table}

Ablations on JFT pre-trained models are shown in Table~\ref{table:ensembles_ablations}; we now discuss the learnings from that.

\begin{table}[tb]
\centering
\caption{\textbf{Ablations}. Test accuracy of different ensembles. 
For each dataset, we take the median of three independent runs. Rows show the average over datasets. 
Bootstrapped confidence intervals at the 95\% level.
Pre-training done on JFT, except for ``All Experts'' that also used ImageNet21k.}
\label{table:ensembles_ablations}
\begin{tabular}{@{}rlcGGG@{}}
\toprule
& Description           & VTAB\textsubscript{1K}  & Natural & Specialised & Structured \\ \midrule
Base & R50 Generalists+Experts           & 76.8\rng{76.4}{77.0}  &  82.6  &  85.8  &  67.2     \\
\textit{Specialists vs} & R50 Experts           & 76.4\rng{76.1}{76.6} & 82.2   & 85.6       & 66.8      \\
\textit{generalists} & R50 Generalists           & 76.5\rng{75.9}{76.7}  &  81.3  &  86.2  &  67.7      \\
\textit{Combine scales} &R18/34/50 Experts     & 76.8\rng{76.4}{77.2} & 82.2  &  85.5       & 67.9  \\ 
\textit{Stack with scale} & R101 Experts      & 77.6\rng{77.4}{77.8} & 83.6   & 86.4        & 68.0      \\
\textit{Massive pool} & All Experts (JFT/INet21k)           & 77.6\rng{77.3}{77.7} & 83.6   & 86.1        & 68.1      \\ 
\bottomrule
\end{tabular}
\end{table}

\textit{Experts help when pre-training is relevant}: Results on Table~\ref{table:upstream_vs_downstream} used $k$NN to pick from a pool of 15 generalists and 244 experts. We break these pools down separately. Experts give a significant boost on Natural datasets; the ensembles take advantage of the relevance of experts pre-trained on the predominately `natural' slices of the upstream data. For datasets without clear experts, there is less benefit to this approach, and the generalists shine. 

\textit{Performance improvements stack with scale}: The strong performance of the R101 Expert Ensemble shows performance gains stack somewhat with scale; it improves on R50 Experts by 1.2\% absolute points in accuracy, improving in all categories. We explore this more thoroughly in Appendix~\ref{app:scale}.

\textit{Combining upstream and downstream diversity helps}:
As discussed in Section \ref{sec:combine}, we combine upstream and downstream diversity. The simple approach of thresholding by $k$NN accuracy works well for choosing the balance between the two at fixed training budget. For R50s, adding augmentations yields a further 0.8\% gain. The number of experts chosen is shown in Appendix \ref{app:num_experts_threshold}.

\subsection{The Value of Nearest Neighbour selection}
\textit{$k$NN may possibly help}:
The greedy ensemble algorithm is not perfect, and with such a small validation set it is prone to overfit.
When \emph{all} upstream JFT R50 experts are fine-tuned and passed to the greedy algorithm, test performance drops slightly. 
We further explore this in Appendices~\ref{app:knn-headroom}, \ref{app:overfitting}.

\textit{It can compare models of different sizes}: Overall, larger models perfom better at transfer \citep{alex2019big}. Per-dataset, this is not the case; e.g. we found R34 experts were best on structured tasks. One may expect $k$NN selection or the greedy algorithm to be biased towards selecting larger architectures. The final ensembles instead use a mix of scales. The R18/R34/R50 experts ensemble improves on just R50s by 0.4\%, indicating possible benefits; more discussion is in Appendix \ref{app:scale}.

\textit{Can filter a very large pool of models}: When selecting only 15 pre-trained models from over 2000 candidates 
(different architecture sizes and upstream datasets), the overall VTAB performance 
(\emph{All Experts} in Table \ref{table:ensembles_ablations} is similar to only selecting from ResNet-101s.
This highlights the remarkable robustness of our model selection.
These results are broken down further in Appendix \ref{app:all}.

Mirroring \citet{puigcerver2020experts}, we have shown $k$NN to be a cheap yet successful way of selecting models. It is not perfect - when combining pools, one would hope for at least a `best of both' performance. 
$k$NN selection wasn't needed for generalists (we had 15 pre-trained models), but when combining the generalists and experts in a pool, specialised/structured performance drops slightly.

%

\subsection{Effect of fine-tuning budget}
\noindent
\begin{minipage}{.588\textwidth}
In most experiments, the $k$NN picks $K = 15$ experts. With the default 4 hyperparameters, this is a fine-tuning budget of 60 models. This is the number of models trained for a given task, and the majority of compute expenditure incurred by a practitioner, as the $k$NN selection/ensembling are comparatively cheap. The hyperensemble was run with the same budget. Figure \ref{fig:finetuning_budget_ensemble_size} shows how performance drops with reduced fine-tuning budget. Interestingly, the expert ensembles are actually more robust to a reduced budget, retaining higher performance when training fewer models, indicating the $k$NN's usefulness as a pre-selection phase.
\end{minipage}\hfill
\begin{minipage}{.39\textwidth}
\centering
\includegraphics[width=0.99\textwidth]{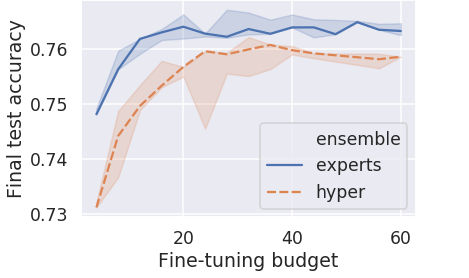}\hspace{-1ex}
\captionof{figure}{Effect of fine-tuning budget on ensemble VTAB\textsubscript{1K} performance.}
\label{fig:finetuning_budget_ensemble_size}
\end{minipage}

\subsection{Robustness to Distribution Shift}
\begin{figure}
\floatbox[{\capbeside\thisfloatsetup{capbesideposition={right,top},capbesidewidth=0.34\textwidth}}]{figure}[\FBwidth]
{\caption{\textbf{Expert ensembles retain higher accuracy under domain shift}. Aside from the first bar, which shows test accuracy, all other bars correspond to some form of induced distribution shift, either artificially or otherwise. In all but one, we get significant boosts in accuracy compared to the HyperEnsembles.}\label{fig:robustness}}
{\includegraphics[width=0.64\textwidth]{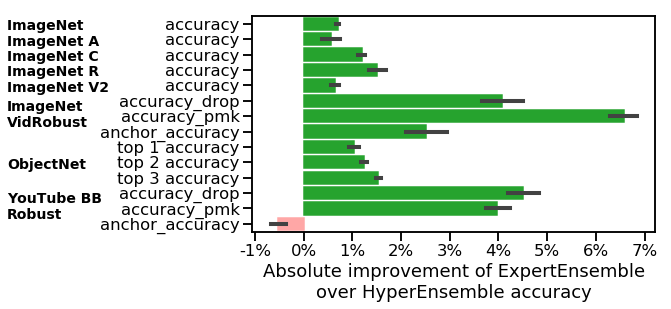}}
\end{figure}

Previous work has shown ensembles help with metrics relating to uncertainty and calibration \citep{balaji2017ensembles,stickl2020diverse}. To assess this, we train JFT R50 HyperEnsembles and ExpertEnsembles for ImageNet classification. For the former, we use the BiT generalist; for the latter, we use the 244 experts, applying $k$NN to 50,000 examples from the training set to select experts. For both we use the validation set for greedy ensembling.
Once the ensembles are trained and constructed, we assess them on a suite of datasets aiming to quantify robustness to distribution shift.
Each dataset introduces some form of distribution shift (further details in Appendix ~\ref{app:imagenet}) - what we assess is the \textit{accuracy} on these datasets.
Figure \ref{fig:robustness} shows them. The expert ensembles offer a slightly better accuracy on the held out data; more importantly, they perform significantly better under distribution shift, improving over the HyperEnsembles by on average 2.2\% across datasets.


\section{Related Work}  
\label{sec:literature}

We present literature related to the main aspects of this work. As well as previous highlighted novelties, we believe our contribution is distinguished from previous ensembling works by focusing on diverse datasets with production-scale deep models (instead of demonstrative smaller architectures and datasets), limiting the training data available, and formally assessing distribution shift.




\textbf{Transfer Learning}.
Relating to a long history of research in manifold and representation learning \citep{chuanqi2018deeptransferlearning}, the transfer of features learnt by deep neural networks aims to reuse the abstract features learnt on a source (upstream) dataset in order to improve performance or data efficiency on a target (downstream) dataset. \citet{bengio2012transfer} studied this in the context of unsupervised pre-training, proposing a number of ways to learn and re-use the features. Many works have shown benefits of transfer relating to convergence speed \citep{raghu2019transfusion}, generalisation \citep{yosinski2014transferable}, accuracy \citep{zhai2019largescale} and robustness \citep{djolonga2020robustness}, with the latter two showing particular benefits in the low-data regime. 

\textbf{Ensemble Learning}.
Known for improving models, ensembling methods have been studied in depth in and out of deep learning academia \citep{seni2010ensemble}.
There are few works which study ensembles in the context of transfer learning \citep{acharya2012transfer, shiliang2013parts}.
\citet{bachman2014pseudo} pre-train entire ensembles on the source data and transfer, instead of transferring individual models.
Work in the low-data regime is sparser. Using an ensemble of models from multiple training checkpoints, \citet{laine2017ssl} label unlabelled data to then train individual models further, improving data efficiency for CIFAR100/SVHN. For few-shot classification on a new class, \citet{dvornik2019fewshot} construct ensembles of mean-centroid classifiers from pre-trained ResNet18s.

\textbf{Deep Ensembles}.
\citet{balaji2017ensembles} show that a simple approach of adversarially training multiple randomly initialised models from scratch and ensembling them yielded models with strong predictive uncertainty and calibration. 
\citet{wenzel2020hyperparameter} showed that hyperensembles, which vary random initialisations \emph{and} hyperparameters, outperform these deep ensembles. 

\textbf{On Constructing Ensembles}.
A key part of our algorithm is the use of the $k$NN to narrow down candidate pre-trained models into a relevant subset. 
\citet{caruana2004ensembleselection} was arguably the seminal work studying how to select an optimal ensemble from a set of candidate models.
A number of works extend AutoML frameworks \citep{he2019automl} to explicitly optimise both the ensembling method and the members to maximise overall performance \citep{wistuba2017frankenstein, xavier2018ea}.


\vspace{-0.5em}
\section{Conclusions}
We have studied simple ways of creating performant ensembles with a limited amount of data.
Overall, ensembles dramatically outperform their single-model counterparts.
We show that diversity from upstream pre-training results in better ensembles than diversity induced downstream, regardless of whether this upstream diversity comes from pre-training multiple generalist models with different initialisations, using different architectures or specialisation via pre-training on different data. We demonstrate the efficacy of the nearest-neighbours classifier as an easily calculated discriminator between different pre-trained models, and even as a way to decide how many models to try on a downstream task, leading to convenient ways to combine both upstream and downstream diversity.

These ensembles achieve SOTA performance on the Visual Task Adaptation Benchmark at a significantly smaller inference cost, while also outperforming ensemble approaches relying on downstream diversity. They also exhibit higher robustness to domain shift as assessed by ImageNet variants.

There are many interesting avenues for future work. All our considered models were pre-trained in a supervised fashion, and this should certainly be extended to include other forms of pre-training. This approach of combining different pre-trained models is highly complimentary with efforts which train ensembles end-to-end with diversity-encouraging losses, such as those in \citet{Lee2015WhyMH} and \citet{webb2019ensemble}. Lastly, works such as Batch Ensembles \citep{wen2020batchensemble} and Parameter Superposition \citep{cheung2019superposition} systematically decompose network parameters to compactly train ensembles. For pre-trained models with the same architecture, weights could be deconstructed to initialise those methods so as to benefit from transfer learning and make them feasible in the low-data regime.

\bibliography{main}

\begin{thebibliography}{54}
\providecommand{\natexlab}[1]{#1}
\providecommand{\url}[1]{\texttt{#1}}
\expandafter\ifx\csname urlstyle\endcsname\relax
  \providecommand{\doi}[1]{doi: #1}\else
  \providecommand{\doi}{doi: \begingroup \urlstyle{rm}\Url}\fi

\bibitem[Acharya et~al.(2012)Acharya, Hruschka, Ghosh, and
  Acharyya]{acharya2012transfer}
Ayan Acharya, Eduardo~R. Hruschka, Joydeep Ghosh, and Sreangsu Acharyya.
\newblock Transfer learning with cluster ensembles.
\newblock In \emph{ICML Workshop on Unsupervised and Transfer Learning}, 2012.

\bibitem[Bachman et~al.(2014)Bachman, Alsharif, and Precup]{bachman2014pseudo}
Philip Bachman, Ouais Alsharif, and Doina Precup.
\newblock Learning with pseudo-ensembles.
\newblock In \emph{Neural Information Processing Systems (NeurIPS)}, 2014.

\bibitem[Barbu et~al.(2019)Barbu, Mayo, Alverio, Luo, Wang, Gutfreund,
  Tenenbaum, and Katz]{objectnet}
Andrei Barbu, David Mayo, Julian Alverio, William Luo, Christopher Wang, Dan
  Gutfreund, Josh Tenenbaum, and Boris Katz.
\newblock Objectnet: A large-scale bias-controlled dataset for pushing the
  limits of object recognition models.
\newblock In \emph{International Conf. on Machine Learning (ICML)}, 2019.

\bibitem[Bengio(2011)]{bengio2012transfer}
Yoshua Bengio.
\newblock Deep learning of representations for unsupervised and transfer
  learning.
\newblock In \emph{ICML Unsupervised and Transfer Learning Workshop}, 2011.

\bibitem[Caruana et~al.(2004)Caruana, Niculescu-Mizil, Crew, and
  Ksikes]{caruana2004ensembleselection}
Rich Caruana, Alexandru Niculescu-Mizil, Geoff Crew, and Alex Ksikes.
\newblock Ensemble selection from libraries of models.
\newblock In \emph{International Conference on Machine Learning (ICML)}, 2004.

\bibitem[Cheng et~al.(2017)Cheng, Han, and Lu]{resisc}
Gong Cheng, Junwei Han, and Xiaoqiang Lu.
\newblock Remote sensing image scene classification: Benchmark and state of the
  art.
\newblock \emph{Proceedings of the IEEE}, 105\penalty0 (10):\penalty0
  1865--1883, Oct 2017.

\bibitem[Cheung et~al.(2019)Cheung, Terekhov, Chen, Agrawal, and
  Olshausen]{cheung2019superposition}
Brian Cheung, Alex Terekhov, Yubei Chen, Pulkit Agrawal, and Bruno Olshausen.
\newblock Superposition of many models into one.
\newblock In \emph{Neural Information Processing Systems (NeurIPS)}, 2019.

\bibitem[Cimpoi et~al.(2014)Cimpoi, Maji, Kokkinos, Mohamed, and Vedaldi]{dtd}
Mircea Cimpoi, Subhransu Maji, Iasonas Kokkinos, Sammy Mohamed, and Andrea
  Vedaldi.
\newblock Describing textures in the wild.
\newblock In \emph{Computer Vision and Pattern Recognition ({CVPR})}, 2014.

\bibitem[Deng et~al.(2009)Deng, Dong, Socher, Li, Li, and Li]{deng2009imagenet}
Jia Deng, Wei Dong, Richard Socher, Li-Jia Li, Kai Li, and Fei-Fei Li.
\newblock {ImageNet}: A large-scale hierarchical image database.
\newblock In \emph{Computer Vision and Pattern Recognition (CVPR)}, 2009.

\bibitem[Dhillon et~al.(2020)Dhillon, Chaudhari, Ravichandran, and
  Soatto]{dhillon2019baseline}
Guneet~S. Dhillon, Pratik Chaudhari, Avinash Ravichandran, and Stefano Soatto.
\newblock A baseline for few-shot image classification.
\newblock In \emph{International Conf. on Learning Representations (ICLR)},
  2020.

\bibitem[Djolonga et~al.(2020)Djolonga, Yung, Tschannen, Romijnders, Beyer,
  Kolesnikov, Puigcerver, Minderer, D'Amour, Moldovan, Gelly, Houlsby, Zhai,
  and Lucic]{djolonga2020robustness}
Josip Djolonga, Jessica Yung, Michael Tschannen, Rob Romijnders, Lucas Beyer,
  Alexander Kolesnikov, Joan Puigcerver, Matthias Minderer, Alexander D'Amour,
  Dan Moldovan, Sylvain Gelly, Neil Houlsby, Xiaohua Zhai, and Mario Lucic.
\newblock On robustness and transferability of convolutional neural networks.
\newblock \emph{arXiv:2007.08558}, 2020.

\bibitem[Dvornik et~al.(2019)Dvornik, Schmid, and Mairal]{dvornik2019fewshot}
Nikita Dvornik, Cordelia Schmid, and Julien Mairal.
\newblock Diversity with cooperation: Ensemble methods for few-shot
  classification.
\newblock In \emph{International Conference on Computer Vision (ICCV)}, 2019.

\bibitem[FAIR(2019)]{pytorch-hub}
FAIR.
\newblock \emph{PyTorch Hub}, 2019.
\newblock URL \url{https://pytorch.org/hub/}.

\bibitem[Fort et~al.(2019)Fort, Hu, and Lakshminarayanan]{fort2019deep}
Stanislav Fort, Huiyi Hu, and Balaji Lakshminarayanan.
\newblock Deep ensembles: A loss landscape perspective.
\newblock \emph{arXiv:1912.02757}, 2019.

\bibitem[Geiger et~al.(2012)Geiger, Lenz, and Urtasun]{kitti}
Andreas Geiger, Philip Lenz, and Raquel Urtasun.
\newblock Are we ready for autonomous driving? the kitti vision benchmark
  suite.
\newblock In \emph{Computer Vision and Pattern Recognition (CVPR)}, 2012.

\bibitem[Google(2018)]{tfhub}
Google.
\newblock \emph{TensorFlow Hub}, 2018.
\newblock URL \url{https://tfhub.dev/}.

\bibitem[He et~al.(2019)He, Zhao, and Chu]{he2019automl}
Xin He, Kaiyong Zhao, and Xiaowen Chu.
\newblock {AutoML}: A survey of the state-of-the-art.
\newblock \emph{arXiv:1908.00709}, 2019.

\bibitem[Helber et~al.(2019)Helber, Bischke, Dengel, and Borth]{eurosat}
Patrick Helber, Benjamin Bischke, Andreas Dengel, and Damian Borth.
\newblock {EuroSAT}: A novel dataset and deep learning benchmark for land use
  and land cover classification.
\newblock \emph{IEEE Journal of Selected Topics in Applied Earth Observations
  and Remote Sensing}, 12\penalty0 (7):\penalty0 2217--2226, 2019.

\bibitem[Hendrycks \& Dietterich(2019)Hendrycks and Dietterich]{imagenet_c}
Dan Hendrycks and Thomas Dietterich.
\newblock Benchmarking neural network robustness to common corruptions and
  perturbations.
\newblock In \emph{International Conf. on Learning Representations (ICLR)},
  2019.

\bibitem[Hendrycks et~al.(2019)Hendrycks, Zhao, Basart, Steinhardt, and
  Song]{imagenet_a}
Dan Hendrycks, Kevin Zhao, Steven Basart, Jacob Steinhardt, and Dawn Song.
\newblock Natural adversarial examples.
\newblock \emph{arXiv:1907.07174}, 2019.

\bibitem[Hendrycks et~al.(2020)Hendrycks, Basart, Mu, Kadavath, Wang, Dorundo,
  Desai, Zhu, Parajuli, Guo, Song, Steinhardt, and Gilmer]{imagenet_r}
Dan Hendrycks, Steven Basart, Norman Mu, Saurav Kadavath, Frank Wang, Evan
  Dorundo, Rahul Desai, Tyler Zhu, Samyak Parajuli, Mike Guo, Dawn Song, Jacob
  Steinhardt, and Justin Gilmer.
\newblock The many faces of robustness: A critical analysis of
  out-of-distribution generalization.
\newblock \emph{arXiv:2006.16241}, 2020.

\bibitem[Johnson et~al.(2017)Johnson, Hariharan, van~der Maaten, Li,
  Lawrence~Zitnick, and Girshick]{clevr}
Justin Johnson, Bharath Hariharan, Laurens van~der Maaten, Fei-Fei Li,
  C~Lawrence~Zitnick, and Ross Girshick.
\newblock {CLEVR}: A diagnostic dataset for compositional language and
  elementary visual reasoning.
\newblock In \emph{Computer Vision and Pattern Recognition (CVPR)}, 2017.

\bibitem[Kaggle \& EyePacs(2015)Kaggle and EyePacs]{retino}
Kaggle and EyePacs.
\newblock Kaggle diabetic retinopathy detection, 2015.
\newblock URL
  \url{https://www.kaggle.com/c/diabetic-retinopathy-detection/data}.

\bibitem[Kolesnikov et~al.(2019)Kolesnikov, Beyer, Zhai, Puigcerver, Yung,
  Gelly, and Houlsby]{alex2019big}
Alexander Kolesnikov, Lucas Beyer, Xiaohua Zhai, Joan Puigcerver, Jessica Yung,
  Sylvain Gelly, and Neil Houlsby.
\newblock Big transfer ({BiT}): General visual representation learning.
\newblock \emph{arXiv:1912.11370}, 2019.

\bibitem[Krizhevsky(2009)]{cifar100}
Alex Krizhevsky.
\newblock Learning multiple layers of features from tiny images.
\newblock Technical report, University of Toronto, 2009.

\bibitem[Laine \& Aila(2017)Laine and Aila]{laine2017ssl}
Samuli Laine and Timo Aila.
\newblock Temporal ensembling for semi-supervised learning.
\newblock In \emph{International Conference on Learning Representations
  (ICLR)}, 2017.

\bibitem[Lakshminarayanan et~al.(2017)Lakshminarayanan, Pritzel, and
  Blundell]{balaji2017ensembles}
Balaji Lakshminarayanan, Alexander Pritzel, and Charles Blundell.
\newblock Simple and scalable predictive uncertainty estimation using deep
  ensembles.
\newblock In \emph{Neural Information Processing Systems (NeurIPS)}, 2017.

\bibitem[LeCun et~al.(2004)LeCun, Huang, and Bottou]{smallnorb}
Yann LeCun, Fu~Jie Huang, and L{\'e}on Bottou.
\newblock Learning methods for generic object recognition with invariance to
  pose and lighting.
\newblock In \emph{Computer Vision and Pattern Recognition (CVPR)}, 2004.

\bibitem[Lee et~al.(2015)Lee, Purushwalkam, Cogswell, Crandall, and
  Batra]{Lee2015WhyMH}
Stefan Lee, Senthil Purushwalkam, M.~Cogswell, David~J. Crandall, and Dhruv
  Batra.
\newblock Why {M} heads are better than one: Training a diverse ensemble of
  deep networks.
\newblock \emph{arXiv:1511.06314}, 2015.

\bibitem[Li et~al.(2004)Li, Fergus, and Perona]{caltech101}
Fei-Fei Li, Rob Fergus, and Pietro Perona.
\newblock Learning generative visual models from few training examples: An
  incremental {Bayesian} approach tested on 101 object categories.
\newblock \emph{Computer Vision and Pattern Recognition Workshop}, 2004.

\bibitem[Matthey et~al.(2017)Matthey, Higgins, Hassabis, and
  Lerchner]{dsprites}
Loic Matthey, Irina Higgins, Demis Hassabis, and Alexander Lerchner.
\newblock {dSprites}: Disentanglement testing sprites dataset, 2017.
\newblock URL \url{https://github.com/deepmind/dsprites-dataset/}.

\bibitem[Netzer et~al.(2011)Netzer, Wang, Coates, Bissacco, and Ng]{svhn}
Yuval Netzer, Tao Wang, Adam Coates, Alessandro Bissacco, and Andrew~Y. Ng.
\newblock Reading digits in natural images with unsupervised feature learning.
\newblock In \emph{NIPS Workshop on Deep Learning and Unsupervised Feature
  Learning 2011}, 2011.

\bibitem[Neyshabur et~al.(2020)Neyshabur, Sedghi, and
  Zhang]{neyshabur2020transferred}
Behnam Neyshabur, Hanie Sedghi, and Chiyuan Zhang.
\newblock What is being transferred in transfer learning?
\newblock \emph{arXiv:2008.11687}, 2020.

\bibitem[Ngiam et~al.(2018)Ngiam, Peng, Vasudevan, Kornblith, Le, and
  Pang]{ngiam2018domain}
Jiquan Ngiam, Daiyi Peng, Vijay Vasudevan, Simon Kornblith, Quoc~V. Le, and
  Ruoming Pang.
\newblock Domain adaptive transfer learning with specialist models.
\newblock \emph{arXiv:1811.07056}, 2018.

\bibitem[Nilsback \& Zisserman(2006)Nilsback and Zisserman]{flowers}
Maria-Elena Nilsback and Andrew Zisserman.
\newblock A visual vocabulary for flower classification.
\newblock In \emph{Computer Vision and Pattern Recognition (CVPR)}, 2006.

\bibitem[Parkhi et~al.(2012)Parkhi, Vedaldi, Zisserman, and Jawahar]{pets}
Omkar~M. Parkhi, Andrea Vedaldi, Andrew Zisserman, and C.~V. Jawahar.
\newblock Cats and dogs.
\newblock In \emph{Computer Vision and Pattern Recognition (CVPR)}, 2012.

\bibitem[Puigcerver et~al.(2020)Puigcerver, Riquelme, Mustafa, Renggli, Pinto,
  Gelly, Keysers, and Houlsby]{puigcerver2020experts}
Joan Puigcerver, Carlos Riquelme, Basil Mustafa, Cedric Renggli, André~Susano
  Pinto, Sylvain Gelly, Daniel Keysers, and Neil Houlsby.
\newblock Scalable transfer learning with expert models.
\newblock \emph{arXiv:2009.13239}, 2020.

\bibitem[Raghu et~al.(2019)Raghu, Zhang, Kleinberg, and
  Bengio]{raghu2019transfusion}
Maithra Raghu, Chiyuan Zhang, Jon Kleinberg, and Samy Bengio.
\newblock Transfusion: Understanding transfer learning for medical imaging.
\newblock In \emph{Neural Information Processing Systems (NeurIPS)}, 2019.

\bibitem[Real et~al.(2017)Real, Shlens, Mazzocchi, Pan, and Vanhoucke]{ytbb}
Esteban Real, Jonathon Shlens, Stefano Mazzocchi, Xin Pan, and Vincent
  Vanhoucke.
\newblock {YouTube}-{BoundingBoxes}: A large high-precision human-annotated
  data set for object detection in video.
\newblock In \emph{Computer Vision and Pattern Recognition (CVPR)}, 2017.

\bibitem[Recht et~al.(2019)Recht, Roelofs, Schmidt, and Shankar]{imagenet_v2}
Benjamin Recht, Rebecca Roelofs, Ludwig Schmidt, and Vaishaal Shankar.
\newblock Do {ImageNet} classifiers generalize to {ImageNet}?
\newblock In \emph{International Conference on Machine Learning (ICML)}, 2019.

\bibitem[Seni \& Elder(2010)Seni and Elder]{seni2010ensemble}
Giovanni Seni and John~F Elder.
\newblock \emph{Ensemble methods in data mining: improving accuracy through
  combining predictions}.
\newblock Morgan \& Claypool Publishers, 2010.

\bibitem[Shankar et~al.(2019)Shankar, Dave, Roelofs, Ramanan, Recht, and
  Schmidt]{shankar2019image}
Vaishaal Shankar, Achal Dave, Rebecca Roelofs, Deva Ramanan, Benjamin Recht,
  and Ludwig Schmidt.
\newblock Do image classifiers generalize across time?
\newblock In \emph{ICML Workshop on Deep Phenomena}, 2019.

\bibitem[Stickland \& Murray(2020)Stickland and Murray]{stickl2020diverse}
Asa~Cooper Stickland and Iain Murray.
\newblock Diverse ensembles improve calibration.
\newblock \emph{arXiv:2007.04206}, 2020.

\bibitem[{Sun} et~al.(2017){Sun}, {Shrivastava}, {Singh}, and {Gupta}]{jft300m}
Chen {Sun}, Abhinav {Shrivastava}, Saurabh {Singh}, and Abhinav {Gupta}.
\newblock Revisiting unreasonable effectiveness of data in deep learning era.
\newblock In \emph{International Conf. on Computer Vision (ICCV)}, 2017.

\bibitem[Sun et~al.(2013)Sun, Xu, and Yang]{shiliang2013parts}
Shiliang Sun, Zhijie Xu, and Mo~Yang.
\newblock Transfer learning with part-based ensembles.
\newblock In \emph{Multiple Classifier Systems}, 2013.

\bibitem[Tan et~al.(2018)Tan, Sun, Kong, Zhang, Yang, and
  Liu]{chuanqi2018deeptransferlearning}
Chuanqi Tan, Fuchun Sun, Tao Kong, Wenchang Zhang, Chao Yang, and Chunfang Liu.
\newblock A survey on deep transfer learning.
\newblock In \emph{International Conf. on Artificial Neural Networks (ICANN)},
  2018.

\bibitem[Veeling et~al.(2018)Veeling, Linmans, Winkens, Cohen, and
  Welling]{camelyon}
Bastiaan~S. Veeling, Jasper Linmans, Jim Winkens, Taco Cohen, and Max Welling.
\newblock Rotation equivariant {CNNs} for digital pathology.
\newblock In \emph{Medical Image Computing and Computer Assisted Intervention
  (MICCAI)}, 2018.

\bibitem[Webb et~al.(2019)Webb, Reynolds, Chen, Reeve, Iliescu, Lujan, and
  Brown]{webb2019ensemble}
Andrew~M. Webb, Charles Reynolds, Wenlin Chen, Henry Reeve, Dan-Andrei Iliescu,
  Mikel Lujan, and Gavin Brown.
\newblock To ensemble or not ensemble: When does end-to-end training fail?
\newblock In \emph{Computer Vision and Pattern Recognition (CVPR)}, 2019.

\bibitem[Wen et~al.(2020)Wen, Tran, and Ba]{wen2020batchensemble}
Yeming Wen, Dustin Tran, and Jimmy Ba.
\newblock Batchensemble: An alternative approach to efficient ensemble and
  lifelong learning.
\newblock In \emph{International Conf. on Learning Representations (ICLR)},
  2020.

\bibitem[Wenzel et~al.(2020)Wenzel, Snoek, Tran, and
  Jenatton]{wenzel2020hyperparameter}
Florian Wenzel, Jasper Snoek, Dustin Tran, and Rodolphe Jenatton.
\newblock Hyperparameter ensembles for robustness and uncertainty
  quantification.
\newblock \emph{arXiv:2006.13570}, 2020.

\bibitem[Wistuba et~al.(2017)Wistuba, Schilling, and
  Schmidt-Thieme]{wistuba2017frankenstein}
Martin Wistuba, Nicolas Schilling, and Lars Schmidt-Thieme.
\newblock Automatic {Frankensteining}: Creating complex ensembles autonomously.
\newblock In \emph{International Conference on Data Mining}, 2017.

\bibitem[{Xavier-Júnior} et~al.(2018){Xavier-Júnior}, {Freitas},
  {Feitosa-Neto}, and {Ludermir}]{xavier2018ea}
João~C. {Xavier-Júnior}, Alex~A. {Freitas}, Antonio {Feitosa-Neto}, and
  Teresa~B. {Ludermir}.
\newblock A novel evolutionary algorithm for automated machine learning
  focusing on classifier ensembles.
\newblock In \emph{Brazilian Conference on Intelligent Systems (BRACIS)}, 2018.

\bibitem[Yosinski et~al.(2014)Yosinski, Clune, Bengio, and
  Lipson]{yosinski2014transferable}
Jason Yosinski, Jeff Clune, Yoshua Bengio, and Hod Lipson.
\newblock How transferable are features in deep neural networks?
\newblock In \emph{Neural information processing systems (NeurIPS)}, 2014.

\bibitem[Zhai et~al.(2019)Zhai, Puigcerver, Kolesnikov, Ruyssen, Riquelme,
  Lucic, Djolonga, Pinto, Neumann, Dosovitskiy, Beyer, Bachem, Tschannen,
  Michalski, Bousquet, Gelly, and Houlsby]{zhai2019largescale}
Xiaohua Zhai, Joan Puigcerver, Alexander Kolesnikov, Pierre Ruyssen, Carlos
  Riquelme, Mario Lucic, Josip Djolonga, Andre~Susano Pinto, Maxim Neumann,
  Alexey Dosovitskiy, Lucas Beyer, Olivier Bachem, Michael Tschannen, Marcin
  Michalski, Olivier Bousquet, Sylvain Gelly, and Neil Houlsby.
\newblock A large-scale study of representation learning with the visual task
  adaptation benchmark.
\newblock \emph{arXiv:1910.04867}, 2019.

\end{thebibliography}
\bibliographystyle{iclr2021_conference}

\newpage
\appendix

\setcounter{topnumber}{5}
\setcounter{bottomnumber}{5}
\setcounter{totalnumber}{10}
\renewcommand{\topfraction}{0.9}
\renewcommand{\bottomfraction}{0.9}
\renewcommand{\textfraction}{0.1}
\renewcommand{\floatpagefraction}{0.9}

\section{Training details}
\subsection{Pretraining pre-trained models}
\label{app:training_fes}
\textbf{Architectures}: The ResNet architectures used is \textit{ResNetV2}, with batch norm replaced by group norm and weight standardisation used in the convolutions, as in \citet{alex2019big}. We consider 18, 34, 50 and 101 layer models.\\ 
\textbf{Datasets}: As mentioned, we use JFT-300M and ImageNet21k as our pretraining datasets. We split these datasets into subsets following the protocol set in \citet{puigcerver2020experts}, where a generalist model is finetuned for 2 epochs on each subset of the pretraining dataset. We repeat this for 18, 34 and 101 layer ResNets on JFT.\\
\textbf{Training}: For the vast majority of pre-trained models in this work, models were pre-trained by taking a generalist model and fine-tuning it for 2 epochs on a subset of the pretraining data. We also consider training architectures from scratch on subsets of the pre-training data; these generally performed slightly worse, but are included in the `All Experts' pool. We pre-train models with the same settings described in \citet{puigcerver2020experts}.

\subsubsection{VTAB: Finetuning pre-trained models}
\label{app:finetuning}
For both generalist and expert models, pre-trained models are trained on a target dataset by adding a new dense layer at the head, with units equal to the number of classes of the output space. All models were trained using SGD with momentum, with the momentum parameter set to 0.99. They were trained on Google Cloud TPUv3s with batch size 512.\\

\textbf{Preprocessing}\\
We use the task-specific preprocessing suggested by BiT-HyperRule \citep{alex2019big}; this dictates resolutions (pre and post crop) and whether to flip left/right. Except for AugEnsembles, we do not use MixUp as they do. For all methods, we train 3 models on each task in order to give rough confidence intervals. As we explore training large numbers of models, it was not computationally feasible to train more.

\textbf{Default Hyperparameter Sweep}\\
Unless hyperparameters are sampled as in Appendix \ref{app:training_hes}, models undergo a default sweep of parameters on the target dataset. For ExpertEnsembles, this means each feature extractor was fine-tuned 4 times on the downstream dataset. For AugEnsembles, it means for a given randomly sampled augmentation setting, the generalist model was fine-tuned 4 times on the downstream dataset. This default sweep is the product of two learning rates (0.1 and 0.01) and two schedule lengths (2500 and 10000 steps). For the schedule, we use an exponential step decay schedule; it warms-up linearly for 20\% of the schedule, then decays by a factor of 0.1 at each subsequent 20\% interval.

\subsection{Training HyperEnsembles}
\label{app:training_hes}
As discussed, our hyper ensembles start with a single pre-trained generalist model - in this case, either a BiT-L or BiT-M ResNet-50x1, which were pretrained on JFT and ImageNet21k respectively. We use the random hyperparameter search space defined in \cite{alex2019big}, which we recount here for convenience:
\begin{itemize}
    \item Weight decay to init $\sim$ \texttt{LogUniform}(1\e{-6}, 1\e{-1})
    \item Dropout rate $\sim$  \texttt{Uniform}($0.0$, $0.5$)
    \item Learning rate $\sim$ \texttt{LogUniform}(1\e{-4}, 1\e{-1})
    \item Schedule length $\sim$ \texttt{Choice}(500, 1k, 2k, 5k, 8k, 16k)
\end{itemize}

\subsection{Training AugEnsembles}
\label{app:training_aes}
Similar to the hyper-ensembles, the AugEnsembles start with a single pre-trained generalist model. For all augmentations, the final resolution for each dataset is that defined by the BiT-HyperRule in \citep{alex2019big}, and we rescale intensities to have a range of -1.0 to 1.0. Note that augmentations themselves are random from image to image, e.g.\ the color distortion augmentations will apply different magnitudes of distortions (hence why for a fixed augmentation setting, there is still benefit to diversity). We sample whether or not to include these distortions and the parameters of these distortions themselves (e.g. distortion level). We randomly sample augmentations as so:
\begin{itemize}
    \item Choose between inception crop and random crop
    \begin{itemize}
        \item If random crop: Sample initial resolution $\sim$ \texttt{Uniform}($1.05$, $1.30$) $\times$ the final resolution
    \end{itemize}
    \item If flipping left-right is mandated in the BiT-HyperRule, choose to flip with probability 50\%.
    \item (For non artificial datasets), apply each of the following color distortions with probability~50\%:
    \begin{itemize}
        \item \textit{Random additive brightness modulation}.\\
        Adds to all channels some $\delta \sim$ \texttt{Uniform}($-\delta_\textrm{max}$, $-\delta_\textrm{max}$).\\
        We sampled $\delta_\textrm{max} \sim$ \texttt{Uniform}($0.01$, $0.2$)
        \item \textit{Random additive hue modulation}.\\
        Adds to the Hue value in HSV space some value $\delta \sim$ \texttt{Uniform}($-\delta_\textrm{max}$, $-\delta_\textrm{max}$).\\
        We sampled $\delta_\textrm{max} \sim$ \texttt{Uniform}($0.01$, $0.2$)
        \item \textit{Random multiplicative saturation modulation.}\\
        Multiplies Saturation channel in HSV by a factor $s \sim $ \texttt{Uniform}($s_\textrm{min}$, $s_\textrm{max}$). \\ We sampled $s_\textrm{min} \sim$ \texttt{Uniform}($0.25$, $0.75$), $s_\textrm{max} \sim$ \texttt{Uniform}($2\times s_\textrm{min}$, $4\times s_\textrm{min}$)
        \item \textit{Random multiplicative contrast modulation.}\\
        Multiplies per-channel standard deviation by a factor $s \sim $ \texttt{Uniform}($s_\textrm{min}$, $s_\textrm{max}$). \\ We sampled $s_\textrm{min} \sim$ \texttt{Uniform}($0.25$, $0.75$), $s_\textrm{max} \sim$ \texttt{Uniform}($2\times s_\textrm{min}$, $4\times s_\textrm{min}$)
    \end{itemize}
    \item Apply mixup with probability 50\%
    \begin{itemize}
        \item If mixup is applied, sample $\alpha$ parameter $\sim$ \texttt{Uniform}($0.01$, $0.2$)
    \end{itemize}
\end{itemize}
Note that we did not apply color distortion to datasets which were artificially generated, e.g. dSprites, as that results in significantly worse performance. That being said, it is not clear what augmentations to use for such datasets. We found the greatest gains from this method to be on specialised datasets. We suspect this is because the default augmentations specified by the BiT HyperRule are not well suited for such datasets; in particular, we found PatchCamelyon benefited the most from different augmentations.

For both augmentation ensembles and hyper ensembles, we checked that the models finetuned on each dataset had a roughly similar distribution of validation accuracies as expert models in order to ascertain these were fair ways of generating models.

\subsection{ImageNet: Training models}
\label{app:imagenet_training}
\textbf{Data}: We use 96\% of the train data for training individual models and the remaining 4\% for constructing ensembles. We report numbers on the official validation split.\\

\textbf{Preprocessing}:  During training we resize to 512 $\times$ 512 pixel images, then randomly crop to 480 $\times$ 480, and also randomly flip images horizontally. At evaluation time we simply resize images to 480 $\times$ 480 pixel. We train five random trials of each approach.\\

\textbf{Training}: We train on Cloud TPUv3s with a batch size of 512, using SGD with Momentum as for VTAB models, with momentum parameter set to 0.99. For the ExpertEnsembles, we use the two learning rates from the default hyperparameter sweep above (0.1 and 0.01), but now use two longer schedules (10k and 20k steps). For the HyperEnsembles, we use the same hyperparameter search space defined in Appendix~\ref{app:training_hes}, but we explore longer schedules of length 10k, 15k, 20k, 25k and 30k steps.

\section{Evaluation details}

\subsection{The Visual Task Adaptation Benchmark (VTAB)}
\label{app:vtab}
The Visual Task Adaptation Benchmark consists of 19 tasks. They are split into three categories:
\begin{itemize}
    \item \textbf{Natural tasks}
    \small{\textcolor{gray}{CalTech101 \citep{caltech101} $\cdot$ CIFAR100 \citep{cifar100} $\cdot$ Street View House Numbers (SVHN - \citet{svhn}) $\cdot$ Describable Textures (DTD - \citet{dtd}) $\cdot$ Oxford Flowers \citep{flowers} $\cdot$ Oxford Pets \citep{pets}}}
    \normalsize These tasks contain `classical' natural real-world images obtained with a camera.
    \item \textbf{Specialised tasks}
    \small{\textcolor{gray}{EuroSAT \citep{eurosat} $\cdot$ Diabetic Retinopothy \citep{retino} PatchCamelyon \citep{camelyon} $\cdot$ Remote Sensing Image Scene Classification (RESISC - \citet{resisc})
    }}
    \normalsize These are datasets of arguably `natural' images which were captured with specialised photographic equipment.
    \item \textbf{Structured datasets}
    \small{\textcolor{gray}{
    DeepMind Lab (Object distance prediction - \citet{zhai2019largescale}) $\cdot$ SmallNOrb (Azimuth \& Elevation prediction - \citet{smallnorb} CLEVR (Counting \& Distance prediction \citet{clevr} $\cdot$ Kitti (Vehicle distance prediction \citet{kitti}) $\cdot$ dSprites (pixel location \& orientation prediction - \citet{dsprites})
    }}
    \normalsize These assess understanding of scene structure in some way, predominately from synthetic environments. Example tasks include 3D depth estimation and counting.
\end{itemize}

\textbf{Validation Performance}.\\
In order to facilitate in-depth study without over-evaluation on the test data, for some plots in the appendix we explore ensemble behaviour on validation data. In this context, given candidate models which have been finetuned on the 800 datapoints, we require data to a) make ensembling decisions (e.g. for the greedy algorithm) and b) evaluate ensemble performance.
We partition the 200 validation points into an 80/20 split, using 160 points to make ensembling decisions and 40 for evaluation. We use $k$-fold cross validation with k = 15, comparing final mean accuracy across iterations and datasets. We generally found that this was representative; i.e. ranking of models according to this setup was the same as for evaluation on the test set.

\subsection{ImageNet Robustness Variants}
\label{app:imagenet}
Given ensembles trained on ImageNet, we assess models using the suite of ImageNet variants collected in \citet{djolonga2020robustness}. Chiefly speaking, these variants induce some kind of distribution shift on the input images while still retaining the same label space.
\begin{itemize}
    \item \textbf{ImageNet Corrupted} (\textit{ImageNet-C} - \citep{imagenet_c})
    The original ImageNet images, with artificial corruptions such as blur and snow applied.
    \item \textbf{Re-collected data}
    These datasets consist of data that was re-collected and labelled with ImageNet labels. ImageNet trained classifiers usually experience an accuracy decrease on these datasets.
    \begin{itemize}
        \item ImageNet Adversarial (\textit{ImageNet-A} - \citep{imagenet_a})
        Collection of new data which ResNet-50 models failed to classify.
        \item ImageNet-R \citep{imagenet_r}
        Collection of cartoons/art/tatoos/toys etc with original ImageNet labels, which ResNet-50 models failed to classify.
        \item ImageNet-v2 \citep{imagenet_v2}
        Data recollected in a method that closely mimics the original protocol as best as possible.
        \item ObjectNet \citep{objectnet}
        Collection of data with controls for backgrounds, rotation and imaging viewpoint. 
    \end{itemize}
    \item \textbf{Video datasets}
    These datasets consist of videos where the frames are labelled with ImageNet classes. In these dataset, as well as the accuracy of the frames, the \textit{pm-k} metric is introduced, whereby given an anchor frame, a classifier is only considered correct if it classifies the surrounding $2k + 1$ frames correctly.
    \begin{itemize}
        \item \textbf{YouTube Bounding Boxes} (YTBB - \citep{ytbb, shankar2019image}
        \item \textbf{ImageNet Vidrobust} \citep{deng2009imagenet, shankar2019image}
    \end{itemize}
\end{itemize}

\section{Ablations and experimental analysis}


\subsection{From-scratch approaches are not competitive in the low-data regime}
\label{app:from_scratch}
\begin{figure}[!htb]
    \centering
    \includegraphics[width=1.0\textwidth]{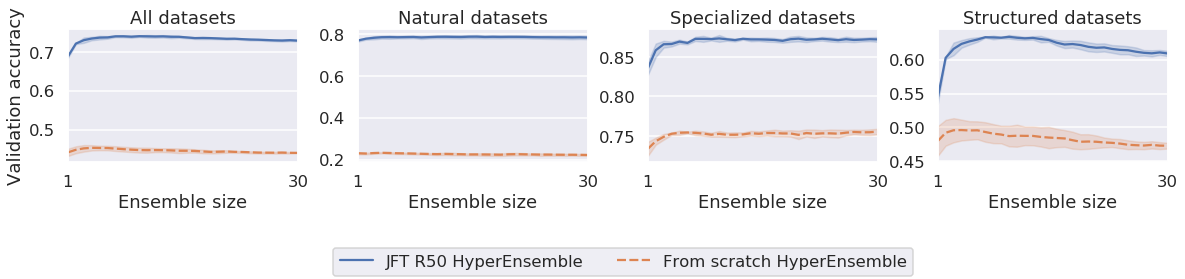} 
    \caption{VTAB\textsubscript{1K} validation accuracy of \emph{HyperEnsembles} trained from scratch and \emph{HyperEnsembles} trained from a generalist pre-trained model (JFT-R50).}
    \label{fig:from_scratch}
\end{figure}

All across the transfer-learning literature there is clear evidence that as the number of available samples for training decreases, the less competitive approaches based on training from scratch become. To demonstrate that also happens when using ensembles we constructed HyperEnsembles starting from a random-initialization and from generalist pre-trained model and evaluate them in VTAK-1K. Figure~\ref{fig:from_scratch} shows the result - training from scratch is hopeless. Even while ensembling many models, it can't compete with a single fine-tuned strong generalist.

\subsection{Number of models being picked}
\label{app:num_models}
Due to the greedy algorithm, different numbers of models are chosen for different datasets. We show in Table \ref{table:num_models} the number of experts chosen for each; in particular we note that on structured datasets, where there are less relevant models from pre-training, it tends to use a large number of models. We note also that typically there is no significant difference in the number of models being picked by different methods.
\begin{table}[b]
\centering
\caption{Median number of experts chosen by each method for each dataset. All are pre-trained on JFT, and all are based on ResNet-50 architectures unless otherwise specified
\label{table:num_models}}
\newcolumntype{R}{@{\hspace{0.078cm}}}
\newcolumntype{K}{>{\raggedleft\arraybackslash}p{0.4cm}}
\begin{tabular}{@{}m{0.07cm}m{2.3cm}>{\raggedleft\arraybackslash}p{0.6cm}|RKRKRKRKRKRKRK|KRKRKRK|@{}KRKRKRKRKRKRKRK@{}}  
\toprule
& &  \rotatebox{90}{mean} & \rotatebox{90}{caltech101} & \rotatebox{90}{cifar100} & \rotatebox{90}{dtd} & \rotatebox{90}{flowers} & \rotatebox{90}{pets} & \rotatebox{90}{sun397} & \rotatebox{90}{svhn} & \rotatebox{90}{camelyon} & \rotatebox{90}{eurosat} & \rotatebox{90}{resisc45} & \rotatebox{90}{retino} & \rotatebox{90}{clever.count} & \rotatebox{90}{clever.closest} & \rotatebox{90}{dmlab} & \rotatebox{90}{dsprites.xpos} & \rotatebox{90}{dsprites.orient} & \rotatebox{90}{kitti} & \rotatebox{90}{smallnorb.azmth} & \rotatebox{90}{smallnorb.elev} \\ 
& & & \multicolumn{7}{c|}{\textbf{Natural}} & \multicolumn{4}{c|}{\textbf{Specialised}} & \multicolumn{8}{c}{\textbf{Structured}}\\ \midrule
D  &  SeedEnsemble  & 9.4 & 5 & 11 & 15 & 2 & 3 & 4 & 10 & 10 & 8 & 13 & 15 & 4 & 15 & 15 & 3 & 15 & 10 & 11 & 15 \\
D  &  AugEnsemble  & 8.2 & 3 & 7 & 13 & 2 & 8 & 4 & 7 & 8 & 5 & 6 & 10 & 3 & 7 & 15 & 10 & 11 & 8 & 15 & 14 \\
D  &  HyperEnsemble  & 9.1 & 10 & 13 & 12 & 2 & 8 & 10 & 10 & 13 & 8 & 8 & 8 & 8 & 11 & 10 & 4 & 7 & 6 & 14 & 12 \\
U  &  Experts  & 9.6 & 4 & 10 & 14 & 1 & 9 & 3 & 11 & 8 & 11 & 12 & 14 & 4 & 14 & 14 & 3 & 13 & 14 & 14 & 14 \\
U  &  Generalists  & 10.9 & 5 & 12 & 15 & 1 & 13 & 14 & 13 & 13 & 9 & 7 & 15 & 5 & 12 & 15 & 3 & 14 & 11 & 15 & 15 \\
U  &  Gen+Experts  & 9.3 & 5 & 8 & 14 & 1 & 6 & 4 & 13 & 8 & 8 & 11 & 14 & 4 & 14 & 14 & 3 & 9 & 14 & 14 & 14 \\
U  &  R18/34/50 Exp.  & 9.1 & 3 & 10 & 14 & 1 & 6 & 3 & 11 & 8 & 7 & 11 & 14 & 4 & 11 & 14 & 2 & 13 & 14 & 14 & 14 \\
U  &  R101 Experts  & 9.7 & 6 & 9 & 14 & 1 & 10 & 5 & 10 & 8 & 8 & 11 & 14 & 6 & 13 & 14 & 2 & 11 & 13 & 14 & 13 \\
C  &  HyperExperts  & 9.9 & 5 & 13 & 15 & 1 & 4 & 7 & 13 & 12 & 9 & 12 & 15 & 2 & 14 & 15 & 2 & 8 & 11 & 15 & 15 \\
C  &  AugExperts  & 10.3 & 4 & 8 & 15 & 1 & 9 & 7 & 15 & 9 & 9 & 8 & 15 & 5 & 15 & 15 & 4 & 11 & 15 & 15 & 15 \\
\bottomrule
\end{tabular}
\end{table}

\subsection{Random vs filtering pre-trained models}
\label{app:knn-vs-random}
\begin{figure}[!htb]
    \centering
    \includegraphics[width=1.0\textwidth]{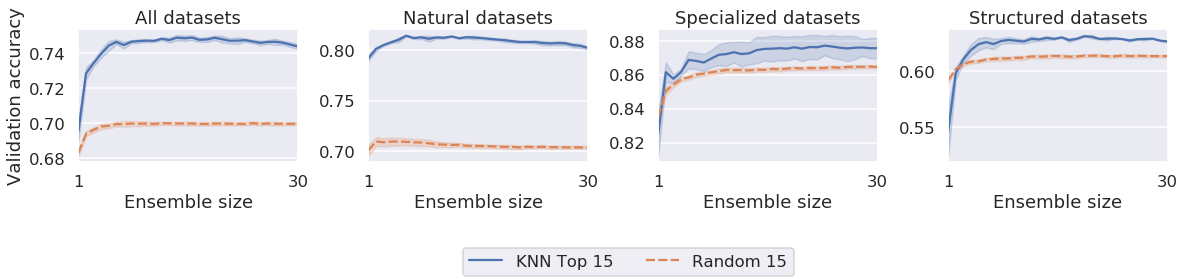} \\
    \caption{Comparison of ensembles created from diversity generated via a random selection of pre-trained models vs using $k$NN to filter pre-trained models. Diverse initialisations alone are not enough; we require individually competitive models.}
    \label{fig:knn-vs-random}
\end{figure}

It is possible that the use of experts, or generalists, is simply the transfer learning equivalent of random initialisation - i.e. that diversity in initialisation for the sake of diversity in initialisation is enough to yield performant models. Figure~\ref{fig:knn-vs-random} shows this is not the case. All categories perform worse, though it is particularly notable in Natural datasets where the ensembles lose the edge of relevant pre-training.

Following from the results of \citet{puigcerver2020experts}, it is clear that the choice of expert is important in the low-data regime. The final results are much more significantly influenced by the initalisation, and transfer from pre-training, than when more data is available. Less suited pre-trained models will lead to lower accuracy and ultimately drag down the predictive power of the entire ensemble, hence why when suitability of pre-trained models varies significantly, the $k$NN becomes much more valuable.

\subsection{Evaluating filtering with \texorpdfstring{$k$NN}{kNN}}
\label{app:knn-headroom}

To evaluate the performance of using $k$NN as a strategy to narrow down a set of pre-trained models, we computed the ensemble test performance on VTAB\textsubscript{1K} using the 244 JFT R50 pre-trained experts with the following strategies:

\textbf{Brute + Greedy Ensemble}: finetune each of the 244 available pre-trained models using the default sweep (4 hyper-parameters). The 976 obtained models are then used to build an ensemble of 15 models using the greedy algorithm.

\textbf{$k$NN top 15 + Greedy Ensemble}: use $k$NN to select 15 out of the 244 available  pre-trained models. The 15 models are finetuned 4 times obtaining 60 models that are then used to build an ensemble of 15 models using the greedy algorithm.

\textbf{$k$NN + Threshold Ensemble}: use $k$NN to select a number ($\leq 15$) of pre-trained models per task using the threshold approach described in Section~\ref{sec:combine}. Each selected models is finetuned 4 times with the default sweep and the one with the best validation accuracy is used in the final ensemble.


The results in Table~\ref{tab:knn-headroom} show that $k$NN filtering was effective at reducing the number of models without drop in VTAB\textsubscript{1K} accuracy. Note also that $k$NN based approaches were able to reduce the noise from the 244 models so that the final ensemble (computed over the validation score) obtained better test accuracy for the natural datasets where we expect to exist clear experts among the models. This conclusion is made stronger by the fact that the \emph{$k$NN + Threshold Ensemble} was also able to beat the brute approach.

\begin{table}[!htb]
\centering
\begin{tabular}{@{}l|c|ccc@{}}
\toprule
                             & VTAB\textsubscript{1K}  & Natural & Specialised & Structured \\ \midrule
Brute + Greedy Ensemble      & 76.20  &  81.17  &  85.46  &  67.22     \\
$k$NN top 15 + Greedy Ensemble  & 76.41  &  82.20  &  85.55  &  66.79     \\ 
$k$NN + Threshold Ensemble         & 76.45  &  82.00  &  85.90  &  66.87     \\ \bottomrule
\end{tabular}
\caption{VTAB\textsubscript{1K} test accuracy of different selection and ensemble strategies. Note that \emph{Brute + Greedy Ensemble} requires finetuning 976 models per task whether the others require 60 or less.}
\label{tab:knn-headroom}
\end{table}

\subsection{Filtering upstream diversity}
\label{app:num_experts_threshold}
\begin{figure}[!htb]
    \centering
    \includegraphics[width=0.5\textwidth]{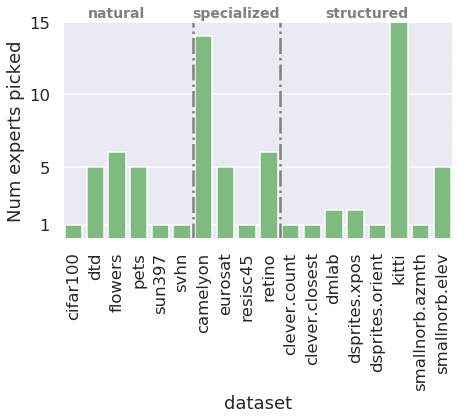}
    \caption{Number of pre-trained models picked by thresholding JFT R50 experts with $\tau=98\%$}
    \label{fig:threshold}
\end{figure}

We presented the use of a simple heuristic to decide how many pre-trained models we should fine-tune on a given task. The remaining fine-tuning budget can then be used for a hyperparameter sweep or varying augmentations; we showed that there were significant gains to be had from doing this. Figure~\ref{fig:threshold} shows the number of experts kept for each dataset using a threshold of $\tau=98\%$.

\subsection{Greedy pre-trained model selection}
\label{app:greedy_NN}
\begin{figure}[!htb]
    \centering
    \includegraphics[width=1.0\textwidth]{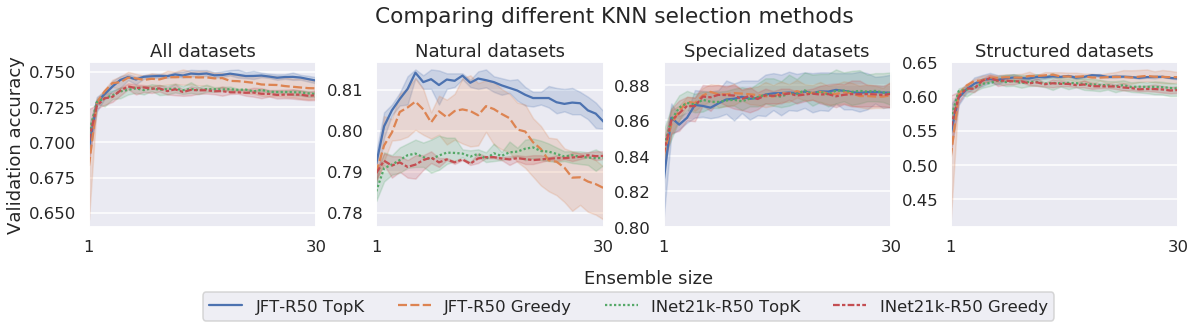}
    \caption{Comparison of greedy $k$NN and top-K $k$NN selection.}
    \label{fig:greedy_vs_topk}
\end{figure}

We experimented with a variant of the $k$NN selection which aimed to pick pre-trained models that would ensemble well together (Greedy), as opposed to picking pre-trained models which are independently accurate (top-K). The effect of this for JFT and ImageNet21k ResNet-50 experts is shown in Figure \ref{fig:greedy_vs_topk}. The greedy approach does not give any clear benefits, and we found at test time it resulted in a small drop in performance. We suspect this is due to the small amount of data the $k$NN is making decisions based off.

\subsection{Selecting from multiple scales of pre-trained models}
\label{app:scale}
\begin{figure}[!htb]
    \centering
    \includegraphics[width=1.0\textwidth]{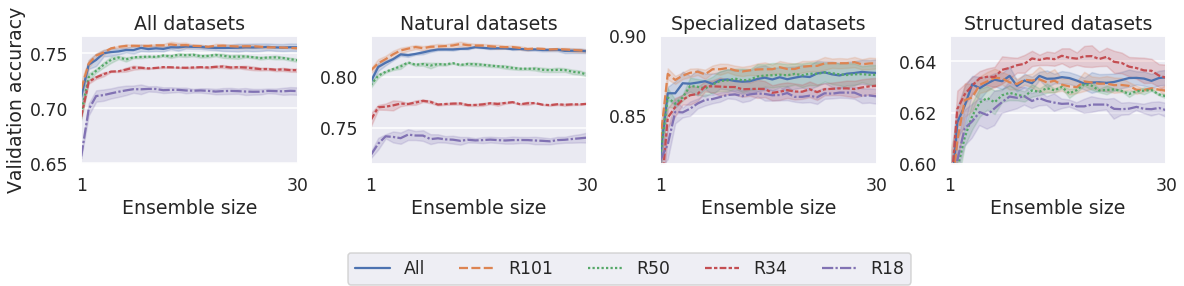}
    \caption{Expert ensembles created from pretraining on JFT subsets at architecture scales. `All' indicates an ExpertEnsemble where the $k$NN selected from all architecture sizes.}
    \label{fig:jft_scales}
\end{figure}

Figure~\ref{fig:jft_scales} shows validation accuracy performance of ensembles of different architecture sizes. Firstly, we note again that to some extent, performance gains stack with architecture size. 
Secondly we note that the larger architectures are not always better; for instance, the ResNet-34s perform significantly better than others on structured datasets. 
Lastly, when considering the `All' line --- where the $k$NN selected from experts pretrained at all four architecture scales -- we would hope it performs as well as the best alternative line. To some extent it does, but it does not always get it right. 

One may expect that the typically higher performing ResNet-101s \citep{alex2019big} would be heavily favoured by the $k$NN algorithm; or failing that, after finetuning on the dataset, would be more performant on the validation set and therefore favoured by the greedy algorithm. Interestingly - on both fronts - this is not the case. The approach is often creating ensembles of diverse architectures scales. We discuss further in Appendix~\ref{app:all}, but it is not clear there is always a clear gain from such diversity. In fact, we see here the $k$NN is not picking as many R34 experts as it should be, despite their clear superiority after fine-tuning; this may be due to comparing features with different dimensions (ResNet-18s and 34s produce 512 dimension features, whereas ResNet-50 and 101 produce 2048 dimensional features), and we suspect using a different distance metric (e.g. cosine distance instead of Euclidean) may have alleviated this.

\begin{figure}
    \centering
    \includegraphics[width=0.9\textwidth]{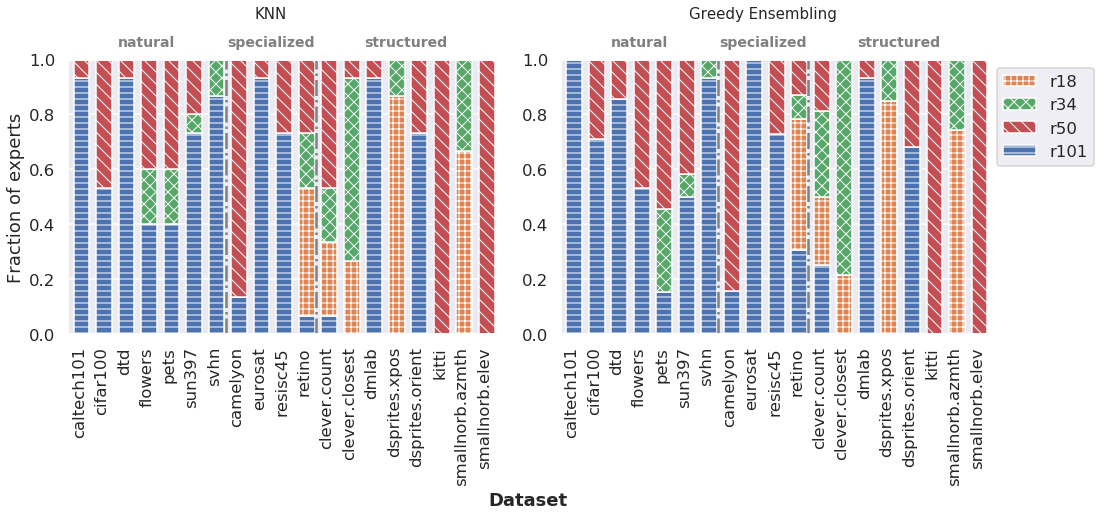}
    \caption{Experts selected when picking from JFT ResNet experts with 18, 34, 50 and 101 layers}
    \label{fig:scales_picked_experts}
\end{figure}



\subsection{Subset selection from a very large pool}
\label{app:all}
\begin{figure}
    \centering
    \includegraphics[width=1.0\textwidth]{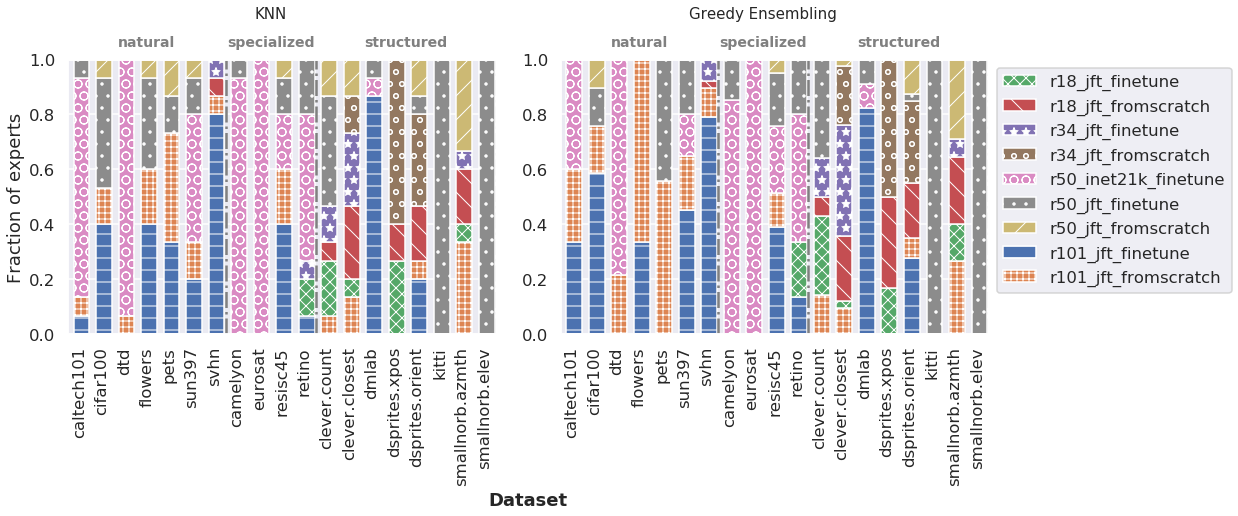}
    \caption{Experts selected when picking from all available feature extractors.}
    \label{fig:all_experts_breakdown}
\end{figure}

\begin{figure}
\floatbox[{\capbeside\thisfloatsetup{capbesideposition={right,top},capbesidewidth=0.3\textwidth}}]{figure}[\FBwidth]
{\caption{Test performance of experts selected from all feature extractors vs.\ those only selected from R101s. Numbers show the number of unique `types' of experts picked; e.g. for \texttt{dtd}, the final ensemble contains ResNet-50s trained on ImageNet21k and ResNet-101s trained on JFT. There is no clear correlation between the number of unique expert types (a form of diversity) and the accuracy.}\label{fig:all_vs_r101}}
{\includegraphics[width=0.68\textwidth]{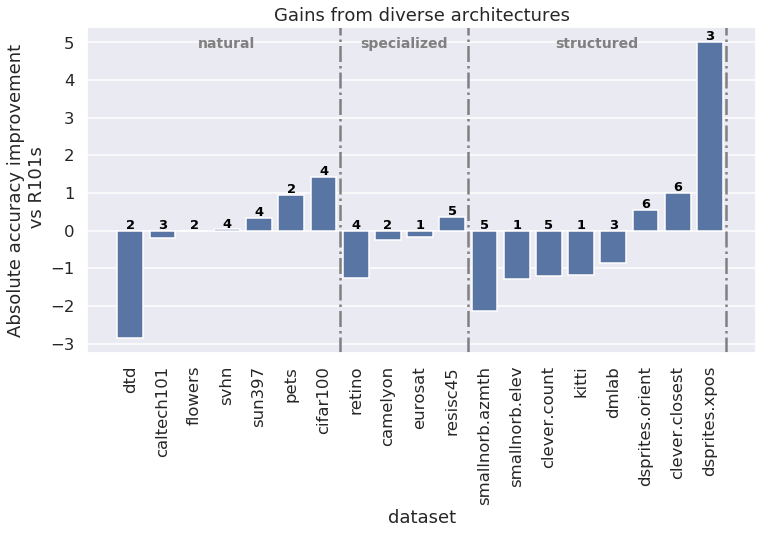}}
\end{figure}

For the majority of our experiments, the $k$NN picked $K = 15$ feature extractors from a pool of 244 (for JFT experts) or 50 (for ImageNet21k experts). Intuitively, the larger the pool of candidate feature extractors is, the harder the job for the $k$NN; and it is clear from the marginal drop in performance when combining the two pools of feature extractors that the task of subset selection will become harder as more candidates are added.

We consider what happens in the extreme limit - as discussed in Appendix \ref{app:training_fes}, as well as the 50 ImageNet21k experts, we trained 4 different architectures of ResNet on 244 different slices of JFT with 2 different pretraining methods (fine-tuning from a generalist and training from scratch). This gives us a total of 2002 experts in our pool.
Figure \ref{fig:all_experts_breakdown} shows the experts picked from this pool. Notably, once again, both the $k$NN and the greedy algorithm pick a diverse selection of models. There are some peculiarities; e.g. for some datasets, ImageNet21k-trained experts perform better as a $k$NN classifier, whereas the JFT experts are better after finetuning, raising issues when using the former as a proxy for the latter.

The interesting question is: are we benefiting from diverse architectures?
We suspect a-priori that the best-performing models will be the R101 experts trained on JFT.
We therefore compare the performance of the diverse expert ensembles with the R101 ones - this is shown in Figure \ref{fig:all_vs_r101}.
It is not immediately clear the existence of any correlation between the diversity of ensembles -- in terms of architecture etc.\ -- with the final test performance. That is not to say there is no theoretical benefit to that at all; this is obfuscated by the extra $k$NN stage, which means there are almost certainly more optimal combinations of different architectures/pretraining styles.

\subsection{Avoiding overfitting}
\label{app:overfitting}

\begin{figure}[tb]
\centering
\begin{subfigure}{0.64\textwidth}
  \centering
  \includegraphics[height=4cm]{images/flops_jft.png}
  \caption{JFT pre-trained models (xent)}
  \label{fig:flops_jft_xent}
\end{subfigure}%
\begin{subfigure}{.33\textwidth}
  \centering
  \includegraphics[height=4cm]{images/flops_inet21k.png}
  \caption{Imagenet21k pre-trained models (xent)}
  \label{fig:flops_inet21k_xent}
\end{subfigure}
\begin{subfigure}{0.64\textwidth}
  \centering
  \includegraphics[height=4cm]{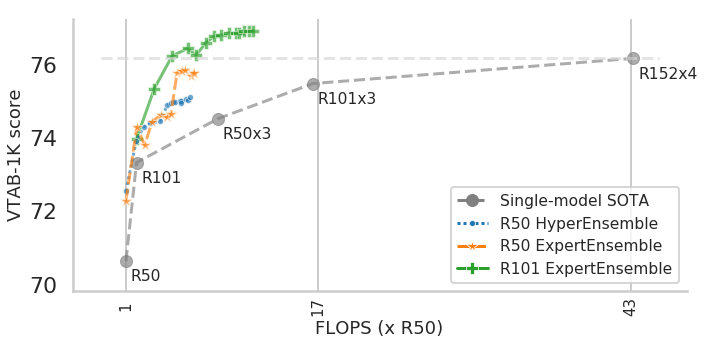}
  \caption{JFT pre-trained models (acc)}
  \label{fig:flops_jft_acc}
\end{subfigure}%
\begin{subfigure}{.33\textwidth}
  \centering
  \includegraphics[height=4cm]{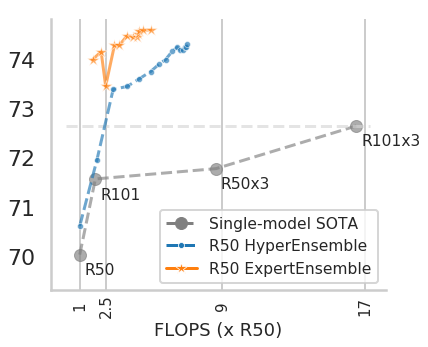}
  \caption{Imagenet21k pre-trained models (acc)}
  \label{fig:flops_inet21k_acc}
\end{subfigure}
\caption{Effect of different greedy ensembling selection strategies. We vary the inference budget to generate ensemble curves. Note that when selecting based on accuracy, the ensembles perform better with smaller ensembles, but that performance doesn't monotonically increase with increasing budget, and overall performance is worse.}
\label{fig:xent_vs_acc}
\end{figure}

All ensembling approaches attempted in this paper experienced higher drops in performance from validation to the test set. It has been previously noted that, especially with small validation datasets, the ensembling process itself can overfit to the validation data \citep{caruana2004ensembleselection}. We attempted a few methods to alleviate this:

\textbf{Bootstrapping data}:
    At each iteration of the greedy algorithm, we randomly bootstrap the 200 validation data points, with some proportion $M$ (such that $200M$ samples are used to make ensembling decisions). This should ensure the ensembling process is not overfitting to individual data points too much, a phenomenon we observed when constructing greedy ensembles based on accuracy.
    This helped marginally when selecting ensembles for maximum accuracy, but we did not see any improvement when using this to select ensembles for maximum cross-entropy.
    
\textbf{Greedy selection with replacement}:
    \citet{caruana2004ensembleselection} reported that allowing the algorithm to re-select models it had already selected reduced overfitting. The extra benefit (in our view, which is not discussed in that work) is that by weighting ensembles according to the number of times they were picked, we can use a weighted average, downweighting less performant models and potentially improving performance. In our set up unfortunately, neither of these held; it did not reduce overfitting, and weighting did not improve performance.
    
\textbf{Ensembling to maximise cross-entropy}:
    We initially constructed ensembles in order to maximise accuracy on the validation set. This resulted in multiple problems; with so few data points, ensembles were making decisions based on single data points (fighting for 0.5\% improvement in accuracies on the 200 data point validation set). Furthermore, multiple ensembles generated by the greedy process would give the exact same validation accuracy, resulting in the need for an arbitrary tiebreaker. For all model sets, with upstream or downstream diversity, maximising cross entropy generalised better and avoided the latter problem, and was the most successful approach to avoiding over-fitting. As well as overall improving performance, we found it more reliable - when making ensembling decisions based on cross-entropy, final test performance monotonically increased with the inference budget (ensemble size). 
    It is worth noting however that decisions made based on cross-entropy are worse for single models or smaller ensembles. With a small inference budget, we suggest optimising for accuracy. These phenomena are evident in Figure \ref{fig:xent_vs_acc}.

Note that \citet{caruana2004ensembleselection} suggest bagging of \textit{ensembles} - i.e. selecting multiple ensembles using different splits of the data, then combining them in one pool. We did not opt for this, as we wanted to have a fixed downstream inference budget, and it is non-trivial to ensure that when combining multiple ensembles of variable size.

\end{document}